\documentclass[11pt,a4paper]{article}
\usepackage[hyperref]{acl2017}
\usepackage[utf8]{inputenc}

\usepackage{times}
\usepackage{latexsym}
\usepackage{graphicx}
\usepackage{multirow}

\usepackage{caption}
\usepackage{subcaption}
\usepackage{booktabs}

\usepackage{color}

\usepackage{url}

\aclfinalcopy 

\newcommand\Mark[1]{\textsuperscript#1}

\title{Enriching Complex Networks with Word Embeddings for Detecting Mild Cognitive Impairment from Speech Transcripts}

\author{
	Leandro B. dos Santos\Mark{1}, Edilson A. Corr{\^e}a Jr\Mark{1}, Osvaldo N. Oliveira Jr\Mark{2}, Diego R. Amancio\Mark{1}, \\ \textbf{Letícia L. Mansur\Mark{3}}, \textbf{Sandra M. Aluísio\Mark{1}}\\
	\Mark{1} Institute of Mathematics and Computer Science, University of S\~{a}o Paulo, S\~{a}o Carlos, S\~{a}o Paulo, Brazil\\
	\Mark{2} S\~{a}o Carlos Institute of Physics, University of S\~{a}o Paulo, S\~{a}o Carlos, S\~{a}o Paulo, Brazil \\
	\Mark{3} Department of Physiotherapy, Speech Pathology and Occupational Therapy, \\ University of S\~{a}o Paulo, S\~{a}o Paulo, S\~{a}o Paulo, Brazil\\
	{\tt \{leandrobs,edilsonacjr,lamansur\}@usp.br}, {\tt chu@ifsc.usp.br}\\
	{\tt \{diego,sandra\}@icmc.usp.br}
}

\date{}

\begin{document}
\maketitle
	
\begin{abstract}
	Mild Cognitive Impairment (MCI) is a mental disorder difficult to diagnose. Linguistic features, mainly from parsers, have been used to detect MCI, but this is not suitable for large-scale assessments. MCI disfluencies produce non-grammatical speech that requires manual or high precision automatic correction of transcripts.  In this paper, we modeled transcripts into complex networks and enriched them with word embedding (CNE) to better represent short texts produced in neuropsychological assessments. The network measurements were applied with well-known classifiers to automatically identify MCI in transcripts, in a binary classification task. A comparison was made with the performance of traditional approaches using Bag of Words (BoW) and linguistic features for three datasets: DementiaBank in English, and Cinderella and Arizona-Battery in Portuguese. Overall, CNE provided higher accuracy than using only complex networks, while Support Vector Machine was superior to other classifiers. CNE provided the highest accuracies for DementiaBank and Cinderella, but BoW was more efficient for the Arizona-Battery dataset probably owing to its short narratives. The approach using linguistic features yielded higher accuracy if the transcriptions of the Cinderella dataset were manually revised. Taken together, the results indicate that complex networks enriched with embedding is promising for detecting MCI in large-scale assessments.
\end{abstract}
	
\section{Introduction\label{sec:intro}}

Mild Cognitive Impairment (MCI) can affect one or multiple cognitive domains (e.g. memory, language, visuospatial skills and executive functions), and may represent a pre-clinical stage of Alzheimer's disease (AD). The impairment that affects memory, referred to as amnestic MCI, is the most frequent, with the highest conversion rate for AD, at 15\% per year versus 1 to 2\% for the general population. Since dementias are chronic and progressive diseases, their early diagnosis ensures a greater chance of success to engage patients in non-pharmacological treatment strategies such as cognitive training, physical activity and socialization \citep{Art:Teixeira:2012:Non:pharmacological}.

Language is one of the most efficient information sources to assess cognitive functions. Changes in language usage are frequent in patients with dementia and are normally first recognized by the patients themselves or their family members. Therefore, the automatic analysis of discourse production is promising in diagnosing MCI at early stages, which may address potentially reversible factors \citep{Art:Muangpaisan:2012:Prevalence}. Proposals to detect language-related impairment in dementias include machine learning \citep{Inc:Jarrold:2010:Language,Art:Roark:2011:Spoken,Art:Fraser:2014:Automated,Art:Fraser:2015:linguistic},  magnetic resonance imaging  \citep{Artc:Dyrba:2015:Predicting}, and data screening tests added to demographic information \citep{Art:Weakley:2015:Neuropsychological}. Discourse production (mainly narratives) is attractive because it allows the analysis of linguistic microstructures, including phonetic-phonological, morphosyntactic and semantic-lexical components, as well as semantic-pragmatic macrostructures.

Automated discourse analysis based on Natural Language Processing (NLP) resources and tools to diagnose dementias via machine learning methods has been used for English language \citep{Inp:Lehr:2012:Fully,Inp:Jarrold:2014:Aided,Inp:Orimaye:2014:Learning,Art:Fraser:2015:linguistic,Inp:Davy:2016:Towards} and for Brazilian Portuguese \citep{Inp:Aluisio:2016:Evaluating}. A variety of features are required for this analysis, including Part-of-Speech (PoS), syntactic complexity, lexical diversity and acoustic features. Producing robust tools to extract these features is extremely difficult because speech transcripts used in neuropsychological evaluations contain disfluencies (repetitions, revisions, paraphasias) and patient's comments about the task being evaluated. Another problem in using linguistic knowledge is the high dependence on manually created resources, such as hand-crafted linguistic rules and/or annotated corpora. Even when traditional statistical techniques (Bag of Words or ngrams) are applied, problems still appear in dealing with disfluencies, because mispronounced words will not be counted together. Indeed, other types of disfluencies (repetition, amendments, patient's comments about the task) will be counted, thus increasing the vocabulary.

An approach applied successfully to several areas of NLP~\citep{Boo:Mihalcea:2011:Graph:NLP}, which may suffer less from the problems mentioned above, relies on the use of complex networks and graph theory.  The word adjacency network model \citep{Art:Cancho:2001:Small,Art:Roxas:2010:prose:poetry,Art:Amancio:2012:Extractive,Art:Amancio:2015:Complex} has provided good results in text classification~\citep{Art:Arruda:2016:Classification:Texts} and related tasks, namely author detection~\citep{Art:Amancio:2015:Authorship}, identification of literary movements~\citep{Art:Amancio:2012:Literary:movements}, authenticity verification~\citep{10.1371/journal.pone.0067310} and word sense discrimination~\citep{0295-5075-98-1-18002}.

In this paper, we show that speech transcripts (narratives or descriptions) can be modeled into complex networks that are enriched with word embedding in order to better represent short texts produced in these assessments. When applied to a machine learning classifier, the complex network features were able to distinguish between control participants and mild cognitive impairment participants. Discrimination of the two classes could be improved by combining complex networks with linguistic and traditional statistical features.

With regard to the task of detecting MCI from transcripts, this paper is, to the best of our knowledge, the first to: a) show that classifiers using features extracted from transcripts modeled into complex networks enriched with word embedding present higher accuracy than using only complex networks for 3 datasets; and b) show that for languages that do not have competitive dependency and constituency parsers to exploit syntactic features, e.g. Brazilian Portuguese, complex networks enriched with  word  embedding constitute a source to extract new, language independent features from transcripts. 

\section{Related Work\label{sec:related}}

Detection of memory impairment has been based on linguistic, acoustic, and demographic features, in addition to scores of neuropsychological tests. Linguistic and acoustic features were used to automatically detect aphasia \cite{Art:Fraser:2014:Automated}; and AD \cite{Art:Fraser:2015:linguistic} or dementia \cite{Inp:Orimaye:2014:Learning} in the public corpora of DementiaBank\footnote{\url{talkbank.org/DementiaBank/}}. Other studies distinguished different types of dementia \cite{Art:Garrard:2014:ML:WAB,Inp:Jarrold:2014:Aided}, in which speech samples were elicited using the Picnic picture of the Western Aphasia Battery \cite{Book:Kertesz:1982:Western}. \citet{Inp:Davy:2016:Towards} also used the Picnic scene to detect MCI, where the subjects were asked to write (by hand) a detailed description of the scene.

As for automatic detection of MCI in narrative speech, \citet{Art:Roark:2011:Spoken} extracted speech features and linguistic complexity measures of speech samples obtained with the Wechsler Logical Memory (WLM) subtest \cite{Book:Wechsler:1997:WLM}, and \citet{Inp:Lehr:2012:Fully} fully automatized the WLM subtest. In this test, the examiner tells a short narrative to a subject, who then retells the story to the examiner, immediately and after a 30-minute delay. WLM scores are obtained by counting the number of story elements recalled.

\citet{Inp:Toth:2015:Automatic} and \citet{Inp:Vincze:2016:Detecting} used short animated films to evaluate immediate and delayed recalls in MCI patients who were asked to talk about the first film shown, then about their previous day, and finally about another film shown last. \citet{Inp:Toth:2015:Automatic} adopted automatic speech recognition (ASR) to extract a phonetic level segmentation, which was used to calculate acoustic features. \citet{Inp:Vincze:2016:Detecting} used speech, morphological, semantic, and demographic features collected from their speech transcripts to automatically identify patients suffering from MCI. 

For the Portuguese language, machine learning algorithms were used to identify subjects with AD and MCI. \citet{Inp:Aluisio:2016:Evaluating} used a variety of linguistic metrics, such as syntactic complexity, idea density \citep{Inp:Cunha:2015:Automatic}, and text cohesion through latent semantics. NLP tools with high precision are needed to compute these metrics, which is a problem for Portuguese since no robust dependency or constituency parsers exist. Therefore, the transcriptions had to be manually revised; they were segmented into sentences, following a semantic-structural criterion and capitalization was applied. The authors also removed disfluencies and inserted omitted subjects when they were hidden, in order to reduce parsing errors. This process is obviously expensive, which has motivated us to use complex networks in the present study to model transcriptions and avoid a manual preprocessing step.

\section{Modeling and Characterizing Texts as Complex Networks}

The theory and concepts of complex networks have been used in several NLP tasks~\citep{Boo:Mihalcea:2011:Graph:NLP,Cong2014598}, such as text classification~\citep{Art:Arruda:2016:Classification:Texts}, summarization~\citep{Antiqueira2009584,Art:Amancio:2012:Extractive} and word sense disambiguation~\citep{0295-5075-98-5-58001}. In this study, we used the word co-occurrence model (also called word adjacency model) because most of the syntactical relations occur among neighboring words~\citep{Art:Cancho:2004:Patterns}.  Each distinct word becomes a node and words that are adjacent in the text are connected by an edge. Mathematically, a network is defined as an undirected graph $G = \{V, E\}$, formed by a set $V = \{v_1, v_2, ..., v_n\}$ of nodes (words) and a set $E = \{e_1, e_2, ..., e_m\}$ of edges (co-occurrence) that are represented by an adjacency matrix $A$, whose elements $A_{ij}$ are equal to $1$ whenever there is an edge connecting nodes (words) $i$ and $j$, and equal to $0$ otherwise.

Before modeling texts into complex networks, it is often necessary to do some preprocessing in the raw text. Preprocessing starts with tokenization where each document/text is divided into tokens (meaningful elements, e.g., words and punctuation marks) and then \textit{stopwords} and punctuation marks are removed, since they have little semantic meaning. One last step we decided to eliminate from the preprocessing pipeline is lemmatization, which transforms each word into its canonical form.
This decision was made based on two factors. First, a recent work has shown that lemmatization has little or no influence when network modeling is adopted in related tasks~\citep{Art:Machicao:2016:Lemma:Influence}. Second, the lemmatization process requires part-of-speech (POS) tagging that may introduce undesirable noises/errors in the text, since the transcriptions in our work contain disfluencies. 

\begin{figure}
	\centering
	\includegraphics[scale=0.2]{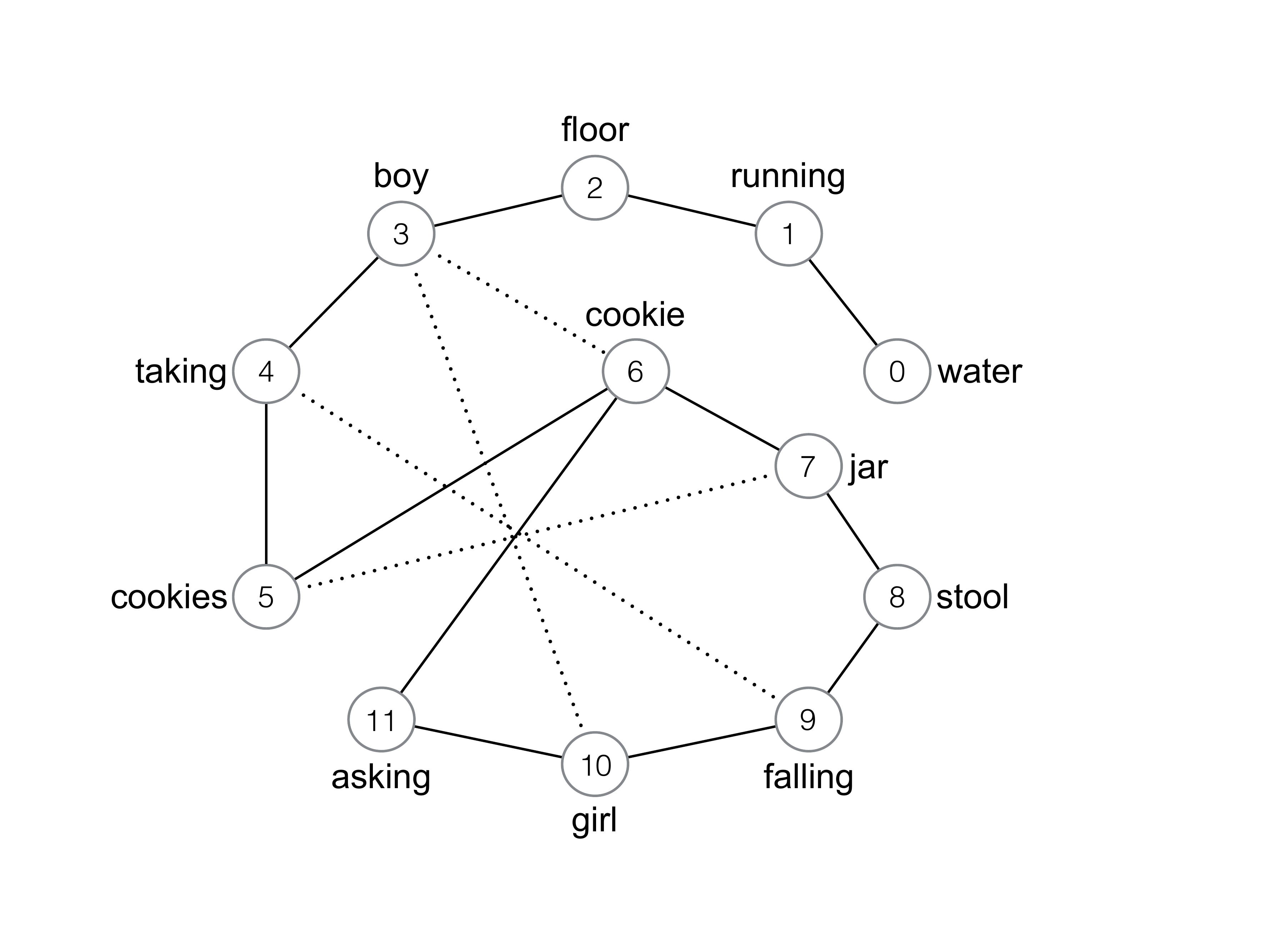}
	\caption{Example of co-occurrence network enriched with semantic information for the following transcription: ``\textit{The water's running on the floor. Boy's taking cookies out of cookie out of the cookie jar. The stool is falling over. The girl was asking for a cookie.}''. The solid edges of the network represent co-occurrence edges and the dotted edges represent connections between words that had similarity higher than $0.5$. \label{fig:complex:network}}
\end{figure}

Another problem with transcriptions in our work is their size. As demonstrated by~\citet{Art:Amancio:2015:Short:Texts}, classification of small texts using networks can be impaired, since short texts have almost linear networks, and the topological measures of these networks have little or no information relevant to classification. To solve this problem, we adapted the approach of inducing language networks from word embeddings, proposed by \citet{Col:Perozzi:2014:Inducing} to enrich the networks with semantic information. In their work, language networks were generated from continuous word representations, in which each word is represented by a dense, real-valued vector obtained by training neural networks in the language model task (or variations, such as context prediction)~\cite{Art:Bengio:2003:Neural,Art:Collobert:2011:NLP,Art:Mikolov:2013:Exploiting,Inp:Mikolov:2013:Distributed}. This structure is known to capture syntactic and semantic information. \citet{Col:Perozzi:2014:Inducing}, in particular, take advantage of word embeddings to build networks where each word is a vertex and edges are defined by similarity between words established by the proximity of the word vectors.

\begin{figure}
	\centering
	\begin{subfigure}[b]{0.22\textwidth}
		\includegraphics[width=\textwidth]{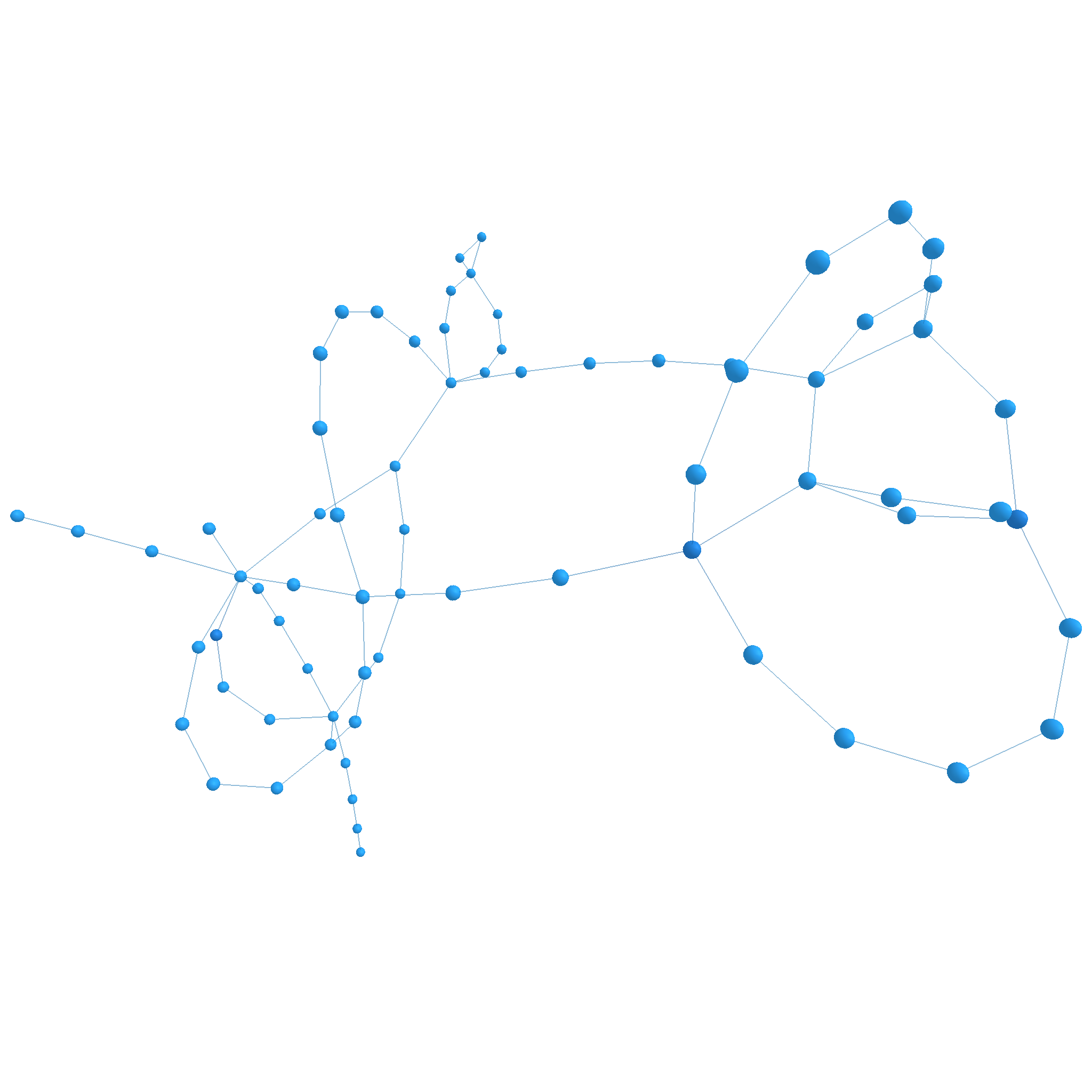}
		\caption{}
		\label{fig:1}
	\end{subfigure}
	\begin{subfigure}[b]{0.22\textwidth}
		\includegraphics[width=\textwidth]{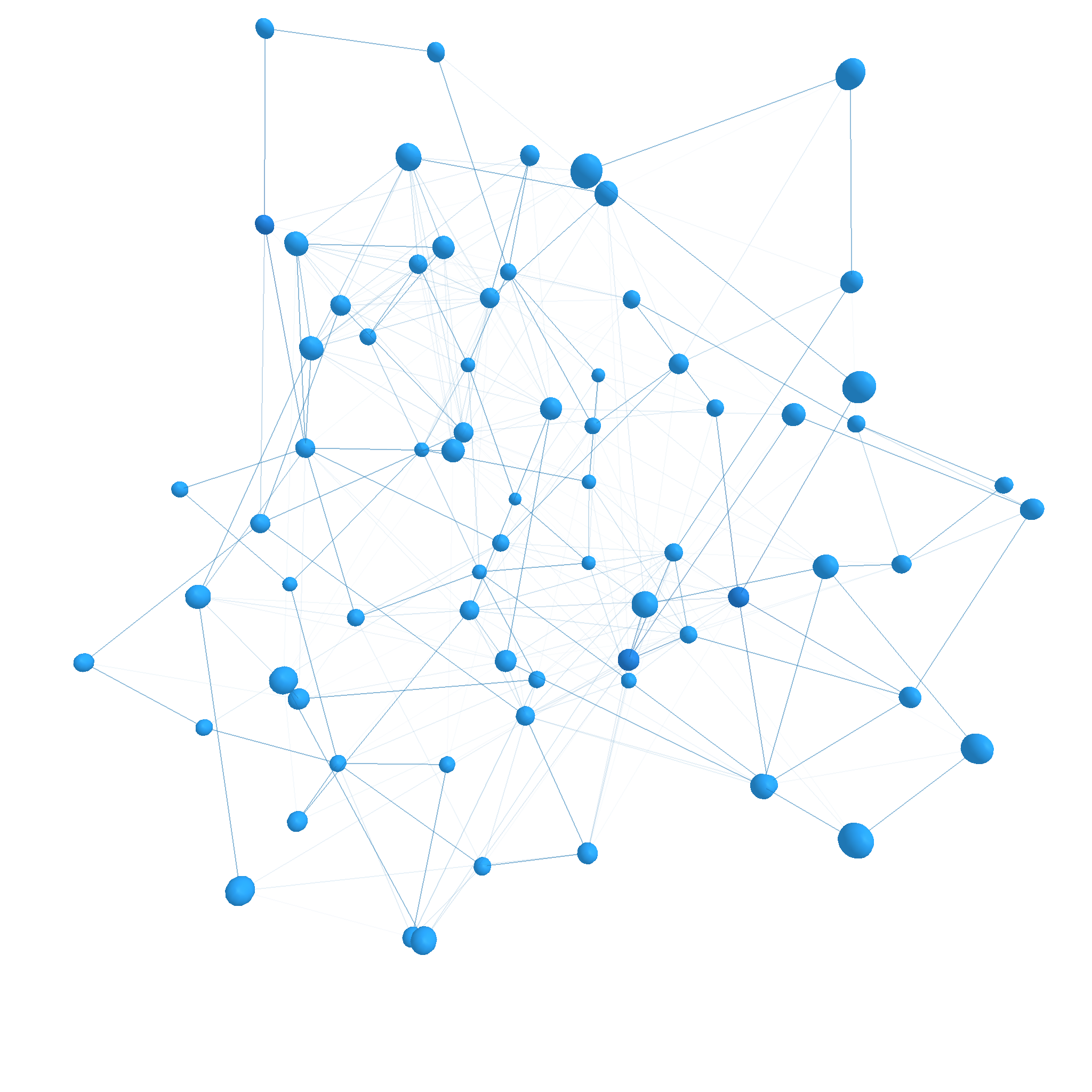}
		\caption{}
		\label{fig:2}
	\end{subfigure}
	\caption{Example of (a) co-occurrence network created for a transcript of the Cookie Theft dataset (see Supplementary Information, Section \ref{sec:supplemental}) and (b) the same co-occurrence network enriched with semantic information. Note that (b) is a more informative network than (a), since (a) is practically a linear network.   \label{fig:complex:network:transcriotion}}
\end{figure}

Following this methodology, in our model we added new edges to the co-occurrence networks considering similarities between words, that is, for all pairs of words in the text that were not connected, an edge was created if their vectors (from word embedding) had a cosine similarity higher than a given threshold. 
Figure~\ref{fig:complex:network} shows an example of a co-occurrence network enriched by similarity links (the dotted edges). 
The gain in information by enriching a co-occurrence network with semantic information is readily apparent in Figure ~\ref{fig:complex:network:transcriotion}.

\section{Datasets, Features and Methods\label{sec:methods}}

\subsection{Datasets}
The datasets\footnote{All datasets are made available in the same representations used in this work, upon request to the authors.} used in our study consisted of: (i) manually segmented and transcribed samples from the DementiaBank and Cinderella story and (ii) transcribed samples of Arizona Battery for Communication Disorders of Dementia (ABCD) automatically segmented into sentences, since we are working towards a fully automated system to detect MCI in transcripts and would like to evaluate a dataset which was automatically processed. 

The DementiaBank dataset is composed of short English descriptions, while the Cinderella dataset contains longer Brazilian Portuguese narratives. ABCD dataset is composed of very short narratives, also in Portuguese. Below, we describe in further detail the datasets, participants, and the task in which they were used.

\subsubsection{The Cookie Theft Picture Description Dataset}

The clinical dataset used for the English language was created during a longitudinal study conducted by the University of Pittsburgh School of Medicine on Alzheimer’s and related dementia, funded by the National Institute of Aging. To be eligible for inclusion in the study, all participants were required to be above 44 years of age, have at least 7 years of education, no history of nervous system disorders nor be taking neuroleptic medication, have an initial Mini-Mental State Exam (MMSE) score of 10 or greater, and be able to give informed consent. The dataset contains transcripts of verbal interviews with AD and related Dementia patients, including those with MCI (for further details see \cite{Art:Becker:1994:Natural}).

We used 43 transcriptions with MCI in addition to another 43 transcriptions sampled from 242 healthy elderly people to be used as the control group. Table \ref{tab:demographic:talkbank} shows the demographic information for the two diagnostic groups.

\begin{table}[!ht]
	\centering
	\small
	\begin{tabular}{c|c|c}
		\hline
		Demographic    & Control           & MCI        \\ \hline
		Avg. Age (SD)  & 64.1 (7.2) & 69.3 (8.2) \\
		No. of Male/Female            & 23/20        & 27/16        \\ \hline
	\end{tabular}
	\caption{Demographic information of participants in the Cookie Theft dataset.}
	\label{tab:demographic:talkbank}
\end{table}

For this dataset, interviews were conducted in English and narrative speech was elicited using the Cookie Theft picture \citep{Book:Goodglass:2001:Assessment} (Figure \ref{fig:book:Cookie:Theft} from \citet{Book:Goodglass:2001:Assessment} in Section \ref{sec:supplemental:example}). During the interview, patients were given the picture and were told to discuss everything they could see happening in the picture. The patients’ verbal utterances were recorded and then transcribed into the CHAT (Codes for the Human Analysis of Transcripts) transcription format \citep{Book:Macwhinney:2000:Childes}.

We extracted the word-level transcript patient sentences from the CHAT files and discarded the annotations, as our goal was to create a fully automated system that does not require the input of a human annotator. We automatically removed filled pauses such as \emph{uh}, \emph{um} , \emph{er} , and \emph{ah}  (e.g. \emph{uh it seems to be summer out}), short false starts (e.g. \emph{just t the ones} ), and repetition (e.g. \emph{mother's finished certain of the the dishes} ), as in \cite{Art:Fraser:2015:linguistic}. The control group had an average of 9.58 sentences per narrative, with each sentence having an average of 9.18 words; while the MCI group had an average of 10.97 sentences per narrative, with 10.33 words per sentence in average.

\subsubsection{The Cinderella Narrative Dataset}

The dataset examined in this study included 20 subjects with MCI and 20 normal elderly control subjects, as diagnosed at the Medical School of the University of S\~ao Paulo (FMUSP). Table \ref{tab:demographic:data} shows the demographic information of the two diagnostic groups, which were also used in \citet{Inp:Aluisio:2016:Evaluating}.

\begin{table}[!ht]
	\small
	\centering
	\begin{tabular}{c|c|c}
		\hline
		Demographic    & Control           & MCI        \\ \hline
		Avg. Age (SD)  & 74.8 (11.3) & 73.3 (5.9) \\
		Avg. Years of  & \multirow{2}{*}{11.4 (2.6)}   & \multirow{2}{*}{10.8 (4.5)} \\
		Education (SD) & & \\
		No. of Male/Female           & 27/16        & 29/14        \\ \hline
	\end{tabular}
	\caption{Demographic information of participants in the Cinderella dataset.}
	\label{tab:demographic:data}
\end{table}

The criteria used to diagnose MCI came from \citet{Art:Petersen:2004:MCI}. Diagnostics were carried out by a  multidisciplinary team consisting of psychiatrists, geriatricians, neurologists, neuropsychologists, speech pathologists, and occupational therapists, by a criterion of consensus. Inclusion criteria  for the control group were elderlies with no cognitive deficits and preservation of functional capacity in everyday life. The exclusion criteria for the normal group were: poorly controlled clinical diseases, sensitive deficits that were not being compensated for and interfered with the performance in tests, and other neurological or psychiatric diagnoses associated with dementia or cognitive deficits and use of medications in doses that affected cognition.

Speech narrative samples were elicited by having participants tell the Cinderella story; participants were given as much time as they needed to examine a picture book illustrating the story (Figure \ref{fig:book:cinderella} in Section \ref{sec:supplemental}). When each participant had finished looking at the pictures, the examiner asked the subject to tell the story in their own words, as in \citet{Art:Saffran:1989:Quantitative}. The time was recorded, but there was no limit imposed to the narrative length. If the participant had difficulty initiating or continuing speech, or took a long pause, an evaluator would use the stimulus question ``What happens next ?'', seeking to encourage the participant to continue his/her narrative. When the subject was unable to proceed with the narrative, the examiner asked if he/she had finished the story and had something to add. Each speech sample was recorded and then manually transcribed at the word level following the  NURC/SP N. 338 EF and 331 D2 transcription norms\footnote{\url{albertofedel.blogspot.com.br/2010_11_01_archive.html}}.

Other tests were applied after the narrative, in the following sequence: phonemic verbal fluency test, action verbal fluency, Camel and Cactus test \citep{Art:Bozeat:2000:Non:Verbal}, and Boston Naming test \citep{Book:Kaplan:2005:Boston}, in order to  diagnose the groups.

Since our ultimate goal is to create a fully automated system that does not require the input of a human annotator, we manually segmented sentences to simulate a high-quality ASR transcript with sentence segmentation, and we automatically removed the disfluencies following the same guidelines of TalkBank project. However, other disfluencies (revisions, elaboration, paraphasias and comments about the task) were kept. The control group had an average of 30.80 sentences per narrative, and each sentence averaged 12.17 words. As for the MCI group, it had an average of 29.90 sentences per narrative, and each sentence averaged 13.03 words.

We also evaluated a different version of the dataset used in \citet{Inp:Aluisio:2016:Evaluating}, where narratives were manually annotated and revised to improve parsing results. The revision process was the following: (i) in the original transcript, segments with hesitations or repetitions of more than one word or segment of a single word were annotated to become a feature and then removed from the narrative to allow the extraction of features from parsing; (ii) empty emissions, which were comments unrelated to the topic of narration or confirmations, such as “\textit{n\'{e}}” (alright), were also annotated and removed; (iii) prolongations of vowels, short pauses and long pauses were also annotated and removed; and (iv) omitted subjects in sentences were inserted. In this revised dataset, the control group had an average of 45.10 sentences per narrative, and each sentence averaged 8.17 words. The MCI group had an average of 31.40 sentences per narrative, with each sentence averaging 10.91 words.

\subsubsection{The ABCD Dataset}

The subtest of immediate/delayed recall of narratives of the ABCD battery was administered to 23 participants with a diagnosis of MCI and 20 normal elderly control participants, as diagnosed at the Medical School of the University of S\~ao Paulo (FMUSP).

MCI subjects produced 46 narratives while the control group produced 39 ones. In order to carry out experiments with a balanced corpus, as with the previous two datasets, we excluded seven transcriptions from the MCI group. We used the automatic sentence segmentation method referred to as DeepBond \cite{Inp:Treviso:2017:DeepBond} in the transcripts.

Table \ref{tab:demographic:data:abcd} shows the demographic information. The control group had an average of 5.23 sentences per narrative, with 11 words per sentence on average, and the MCI group had an average of 4.95 sentences per narrative, with an average of 12.04 words per sentence.  Interviews were conducted in Portuguese and the subject listened to the examiner read a short narrative. The subject then retold the narrative to the examiner twice: once immediately upon hearing it and again after a 30-minute delay \cite{Book:Bayles:1991:ABCD}. Each speech sample was recorded and then manually transcribed at the word level following the NURC/SP N. 338 EF and 331 D2 transcription norms.

\begin{table}[!ht]
	\small
	\centering
	\begin{tabular}{c|c|c}
		\hline
		Demographic    & Control           & MCI        \\ \hline
		Avg. Age (SD)  & 61 (7.5) & 72,0 (7.4) \\
		Avg. Years of  & \multirow{2}{*}{16 (7.6)}   & \multirow{2}{*}{13.3 (4.2)} \\
		Education (SD) & & \\
		No. of Male/Female           & 6/14        & 16/7        \\ \hline
	\end{tabular}
	\caption{Demographic information of participants in the ABCD dataset.}
	\label{tab:demographic:data:abcd}
\end{table}

\subsection{Features}
Features of three distinct natures were used to classify the transcribed texts: topological metrics of co-occurrence networks, linguistic features and bag of words representations.  

\subsubsection{Topological Characterization of Networks}

Each transcription was mapped into a co-occurrence network, and then enriched via word embeddings using the cosine similarity of words. Since the occurrence of out-of-vocabulary words is common in texts of neuropsychological assessments, we used the method proposed by \citet{Art:Bojanowski:2016:Enriching} to generate word embeddings. This method extends the skip-gram model to use character-level information, with each word being represented as a bag of character $n$-grams. It provides some improvement in comparison with the traditional skip-gram model in terms of syntactic evaluation~\cite{Inp:Mikolov:2013:Distributed} but not for semantic evaluation.

Once the network has been enriched, we characterize its topology using the following ten measurements:
\begin{enumerate}
	\item  \textbf{PageRank:} is a centrality measurement that reflects the relevance of a node based on its connections to other relevant nodes \citep{Inp:Brin:2012:PageRank}; 
	\item \textbf{Betweenness:} is a centrality measurement that considers a node as relevant if it is highly accessed via shortest paths. The betweenness of a node $v$ is defined as the fraction of shortest paths going through node $v$;  
	\item \textbf{Eccentricity:} of a node is calculated by measuring the shortest distance from the node to all other vertices in the graph and taking the maximum;   \item \textbf{Eigenvector centrality:} is a measurement that defines the importance of a node based on its connectivity to high-rank nodes;
	\item \textbf{Average Degree of the Neighbors of a Node:} is the average of the degrees of all its direct neighbors; 
	\item \textbf{Average Shortest Path Length of a Node:} is the average distance between this node and all other nodes of the network; 
	\item \textbf{Degree:} is the number of edges connected to the node; 
	\item \textbf{Assortativity Degree:} or degree correlation measures the tendency of nodes to connect to other nodes that have similar degree; 
	\item  \textbf{Diameter:} is defined as the maximum shortest path;  
	\item \textbf{Clustering Coefficient:} measures the probability that two neighbors of a node are connected.
\end{enumerate}

Most of the measurements described above are local measurements, i.e. each node $i$ possesses a value $X_i$, so we calculated the average $\mu(X)$, standard deviation $\sigma(X)$ and skewness $\gamma(X)$ for each measurement~\citep{Art:Amancio:2015:Complex}.

\subsubsection{Linguistic Features}

Linguistic features for classification of neuropsychological assessments have been used in several studies \citep{Art:Roark:2011:Spoken,Inp:Jarrold:2014:Aided,Art:Fraser:2014:Automated,Inp:Orimaye:2014:Learning,Art:Fraser:2015:linguistic,Inp:Vincze:2016:Detecting,Inp:Davy:2016:Towards}. We used the Coh-Metrix\footnote{\url{cohmetrix.com}}\citep{Art:Graesser:2004:Coh:Metrix} tool to extract features from English transcripts,  resulting in 106 features. The metrics are divided into eleven categories: Descriptive, Text Easability  Principal Component, Referential Cohesion, Latent Semantic Analysis  (LSA), Lexical Diversity, Connectives, Situation Model, Syntactic Complexity, Syntactic Pattern Density, Word Information, and Readability (Flesch Reading Ease, Flesch-Kincaid Grade Level, 
Coh-Metrix L2 Readability).

For Portuguese, Coh-Metrix-Dementia \citep{Inp:Aluisio:2016:Evaluating} was used. The metrics affected by constituency and dependency parsing were not used because they are not robust with disfluencies. Metrics based on manual annotation (such as proportion short pauses, mean pause duration, mean number of empty words, and others) were also discarded.  The metrics of Coh-Metrix-Dementia are divided into twelve categories: Ambiguity, Anaphoras, Basic Counts, Connectives, Co-reference Measures, Content Word Frequencies, Hypernyms, Logic Operators, Latent Semantic Analysis, Semantic Density, Syntactical Complexity, and Tokens. The metrics used are shown in detail in Section \ref{sec:supplemental:metrics}. In total, 58 metrics were used, from the 73 available on the website\footnote{\url{http://143.107.183.175:22380}}.

\subsubsection{Bag of Words}

The representation of text collections under the BoW assumption (i.e., with no information relating to word order) has been a robust solution for text classification.
In this methodology, transcripts are represented by a table in which the columns represent the terms (or existing words) in the transcripts and the values represent frequency of a term in a document.

\subsection{Classification Algorithms}

In order to quantify the ability of the topological characterization of networks, linguistic metrics and BoW features were used to distinguish subjects with MCI from healthy controls. We employed four machine learning algorithms to induce classifiers from a training set. These techniques were the Gaussian Naive Bayes (G-NB), $k$-Nearest Neighbor ($k$-NN), Support Vector Machine (SVM), linear and radial bases functions (RBF), and Random Forest (RF). We also combined these classifiers through ensemble and multi-view learning. In ensemble learning, multiple models/classifiers are generated and combined using a majority vote or the average of class probabilities to produce a single result~\citep{Book:Zhou:2012:Ensemble}.

In multi-view learning, multiple classifiers are trained in different feature spaces and thus combined to produce a single result. This approach is an elegant solution in comparison to combining all features in the same vector or space, for two main reasons. First, combination is not a straightforward step and may lead to noise insertion since the data have different natures.  Second, using different classifiers for each feature space allows for different weights to be given for each type of feature, and these weights can be learned by a regression method to improve the model. In this work, we used majority voting to combine different feature spaces.

\section{Experiments and Results\label{sec:experiments}}

All experiments were conducted using the Scikit-learn\footnote{\url{http://scikit-learn.org}} \citep{Art:Pedregosa:scikit-learn:2011}, with classifiers evaluated on the basis of classification accuracy i.e. the total proportion of narratives which were correctly classified. The evaluation was performed using 5-fold cross-validation instead of the well-accepted 10-fold cross-validation because the datasets in our study were small and the test set would have shrunk, leading to less precise measurements of accuracy. The threshold parameter was optimized with the best values being $0.7$ in the Cookie Theft dataset and $0.4$ in both the Cinderella and ABCD datasets. 

We used the model proposed by \citet{Art:Bojanowski:2016:Enriching} with default parameters (100 dimensional embeddings, context window equal to 5 and 5 epochs) to generate word embedding. We trained the models in Portuguese and English Wikipedia dumps from October and November 2016 respectively.

The accuracy in classification is given in Tables \ref{results:en} through \ref{results:pt:abcd}. CN, CNE, LM, and BoW  denote, respectively, complex networks, complex network enriched with embedding, linguistic metrics and Bag of Words, and CNE-LM, CNE-BoW, LM-BoW and CNE-LM-BoW refer to combinations of the feature spaces (multiview learning), using the majority vote. Cells with the ``--'' sign mean that it was not possible to apply majority voting because there were two classifiers. The last line represents the use of an ensemble of machine learning algorithms, in which the combination used was the majority voting in both ensemble and multiview learning.

\begin{table*}[!ht]
	\small
	\centering
	\begin{tabular}{@{}lcccccccc@{}}
		\toprule
		Classifier & CN & CNE & LM & BoW & CNE-LM & CNE-BoW & LM-BoW & CNE-LM-BoW \\ \midrule
		SVM-Linear & 52 & 55 & 56 & 59	&	--	&	--	&	--	&	60 \\
		SVM-RBF & 56 & \textbf{62} & \textbf{58} & \textbf{60}	&	--	&	--	&	--	&	\textbf{65} \\
		$k$-NN & \textbf{59} & 61 & 46 & 57 &	--	&	--	&	--	&	59 \\
		RF & 52 & 47 & 45 & 48	&	--	&	--	&	--	&	50 \\
		G-NB & 51 & 48 & 56 & 55	&	--	&	--	&	--	&	50\\
		Ensemble & 56 & 60 & 54 & 58 & 57 & 60 & 63 & \textbf{65} \\ \bottomrule
	\end{tabular}
	\caption{Classification accuracy achieved on Cookie Theft dataset.}
	\label{results:en}
\end{table*}

\begin{table*}[!ht]
	\small
	\centering
	\begin{tabular}{@{}lcccccccc@{}}
		\toprule
		Classifier & CN & CNE & LM & BoW & CNE-LM & CNE-BoW & LM-BoW & CNE-LM-BoW \\ \midrule
		SVM-Linear & 52 & 60 & \textbf{52} & 50 	&	--	&	--	&	--	& \textbf{52} \\
		SVM-RBF    & \textbf{57} & \textbf{65} & 47 & 37 &--	&	--	&	--	& 50\\
		$k$-NN        & 47 & 50 & 47 & 37	&	--	&	--	&	--	&	37 \\
		RF         & 55 & 57 & 47 & 45 &	--	&	--	&	--	&	\textbf{52} \\
		G-NB       & 47 & 52 & 47 & \textbf{55}	&	--	&	--	&	--	&	\textbf{52}	\\
		Ensemble   & 52 & 60 & 50 & 37 & 57 & 52 & 50 & 47 \\ \bottomrule
	\end{tabular}
	\caption{Classification accuracy achieved on Cinderella dataset.}
	\label{results:pt:automaticremove}
\end{table*}

\begin{table*}[!ht]
	\small
	\centering
	\begin{tabular}{@{}lcccccccccc@{}}
		\toprule
		Classifier & CN & CNE & LM & BoW & CNE-LM & CNE-BoW & LM-BoW & CNE-LM-BoW \\ \midrule
		SVM-Linear & 56 & \textbf{69} & 51 & \textbf{75} & -- & -- & -- & \textbf{74}\\
		SVM-RBF    & 54 & 57 & 66 & 67 & -- & -- & -- & 71\\
		$k$-NN        & 56 & 56 & 69 & 63 & -- & -- & -- & 71\\
		RF	       & 54 & 62 & \textbf{70} & 64 & -- & -- & -- & 69\\
		G-NB       & \textbf{61} & 55 & 55 & 65 & -- & -- & -- & 65\\
		Ensemble   & 55 & 61 & 62 & 72 & 69 & 68 & 75 & 73\\ \bottomrule
	\end{tabular}
	\caption{Classification accuracy achieved on ABCD dataset.}
	\label{results:pt:abcd}
\end{table*}

In general, CNE outperforms the approach using only complex networks (CN), while SVM  (Linear  or  RBF  kernel) provides higher accuracy than other machine learning algorithms. The results for the three datasets show that characterizing transcriptions into complex networks is competitive with other traditional methods, such as the use of linguistic metrics.  In fact, among the three types of features, using enriched networks (CNE) provided the highest accuracies in two datasets (Cookie Theft and original Cinderella). For the ABCD dataset, which contains short narratives, the small length of the transcriptions may have had an effect, since BoW features led to the highest accuracy. In the case of the revised Cinderella dataset, segmented into sentences and capitalized as reported in \citet{Inp:Aluisio:2016:Evaluating}, Table \ref{results:pt:andre} shows that the manual revision was an important factor, since the highest accuracies were obtained with the approach based on linguistic metrics (LM). However, this process of manually removing disfluencies demands time; therefore it is not practical for large-scale assessments.

Ensemble and multi-view learning were helpful for the Cookie Theft dataset, in which multi-view learning achieved the highest accuracy (65\% of accuracy for narrative texts, a 3\% of improvement compared to the best individual classifier). However, neither multi-view or ensemble learning enhanced accuracy in the Cinderella dataset, where SVM-RBF with CNE space achieved the highest accuracy (65\%).  For the ABCD dataset, multi-view CNE-LM-BoW with SVM-RBF and KNN classifiers improved the accuracy to 4\% and 2\%, respectively. Somewhat surprising were the results of SVM with linear kernel in BoW feature space (75\% of accuracy).

\begin{table}[!ht]
	\small
	\centering
	\begin{tabular}{@{}lccccccc@{}}
		\toprule
		Classifier & CN & CNE & LM & BoW  \\ \midrule
		SVM-Linear & 50 & 65 & 65 & 52 \\
		SVM-RBF & \textbf{57} & \textbf{67} & \textbf{72} & \textbf{55} \\
		KNN & 42 & 47 & 55 & 50 \\
		RF & 52 & 47 & 70 & 45 \\
		G-NB  & 52 & 65 & 62 & 45 \\
		Ensemble & 52 & 60 & \textbf{72} & 45 \\ \bottomrule
	\end{tabular}
	\caption{Classification accuracy achieved on Cinderella dataset manually processed to revise non-grammatical sentences.}
	\label{results:pt:andre}
\end{table} 

\section{Conclusions and Future Work \label{sec:conclusion}}

In this study, we employed metrics of topological properties of CN in a machine learning classification approach to distinguish between healthy patients and patients with MCI. To the best of our knowledge, these metrics have never been used to detect MCI in speech transcripts; CN were enriched with word embeddings to better represent short texts produced in neuropsychological assessments. The topological properties of CN outperform traditional linguistic metrics in individual classifiers’ results.  Linguistic features depend on grammatical texts to present good results, as can be seen in the results of the manually processed Cinderella dataset (Table \ref{results:pt:andre}).
Furthermore, we found that combining machine and multi-view learning can improve accuracy. The accuracies found here are comparable to the values reported by other authors, ranging from 60\% to 85\%  \citep{Inp:Prud'hommeaux:2011:Alignment,Inp:Lehr:2012:Fully,Inp:Toth:2015:Automatic,Inp:Vincze:2016:Detecting}, which means that it is not easy to distinguish between healthy subjects and those with cognitive impairments. The comparison with our results is not straightforward, though, because the databases used in the studies are different. There is a clear need for publicly available datasets to compare different methods, which would optimize the detection of MCI in elderly people.

In future work, we intend to explore other methods to enrich CN, such as the Recurrent Language Model, and use other metrics to characterize an adjacency network. The pursuit of these strategies is relevant because language is one of the most efficient information sources to evaluate cognitive functions, commonly used in neuropsychological assessments. As this work is ongoing, we will keep collecting new transcriptions of the ABCD retelling subtest to increase the corpus size and obtain more reliable results in our studies. Our final goal is to apply neuropsychological assessment batteries, such as the ABCD retelling subtest, to mobile devices, specifically tablets. This adaptation will enable large-scale applications in hospitals and facilitate the maintenance of application history in longitudinal studies, by storing the results in databases immediately after the test application. 

\section*{Acknowledgments}

This work was supported by CAPES, CNPq, FAPESP, and Google Research Awards in Latin America. We would like to thank NVIDIA for their donation of GPU.

\bibliography{acl2017}
\bibliographystyle{acl_natbib}

\appendix

\section{Supplementary Material}
\label{sec:supplemental}

\begin{figure}[!ht]
	\centering
	\includegraphics[scale=0.5]{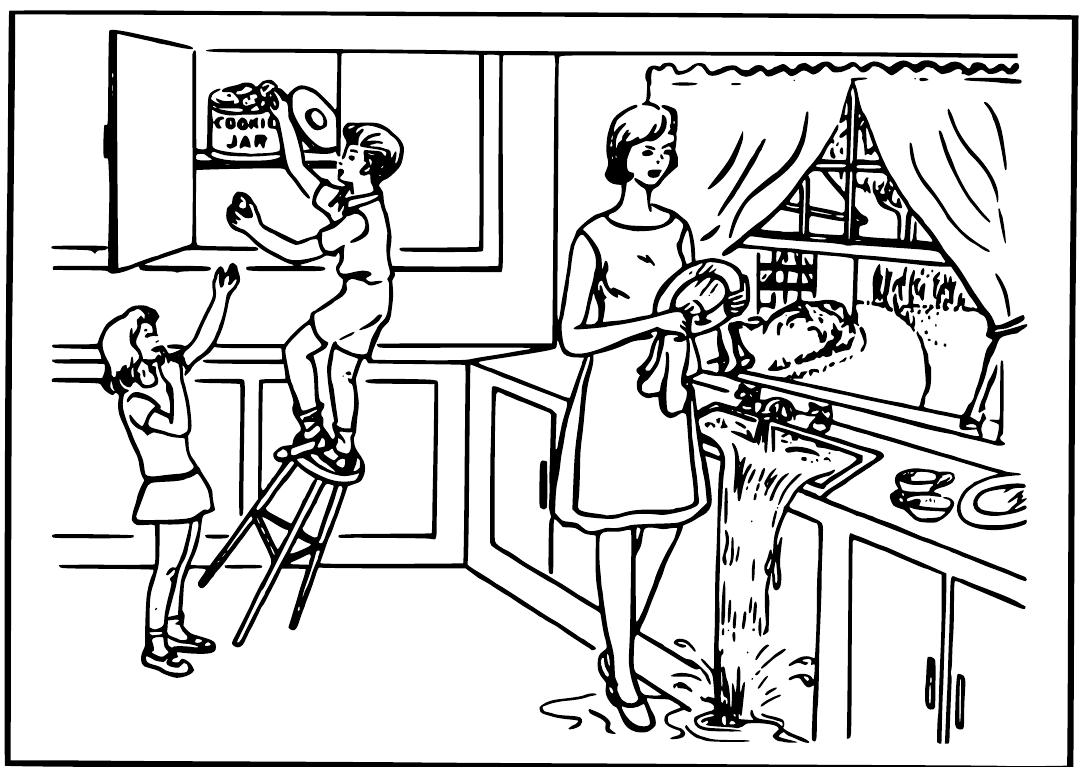}
	\caption{The Cookie Theft Picture, taken from the Boston
		Diagnostic Aphasia Examination \citep{Book:Goodglass:2001:Assessment}.}
	\label{fig:book:Cookie:Theft}
\end{figure}

Figure \ref{fig:book:Cookie:Theft} is Cookie Theft picture, which was used in DementiaBank project.

\begin{figure}[!ht]
	\centering
	\includegraphics[scale=0.6]{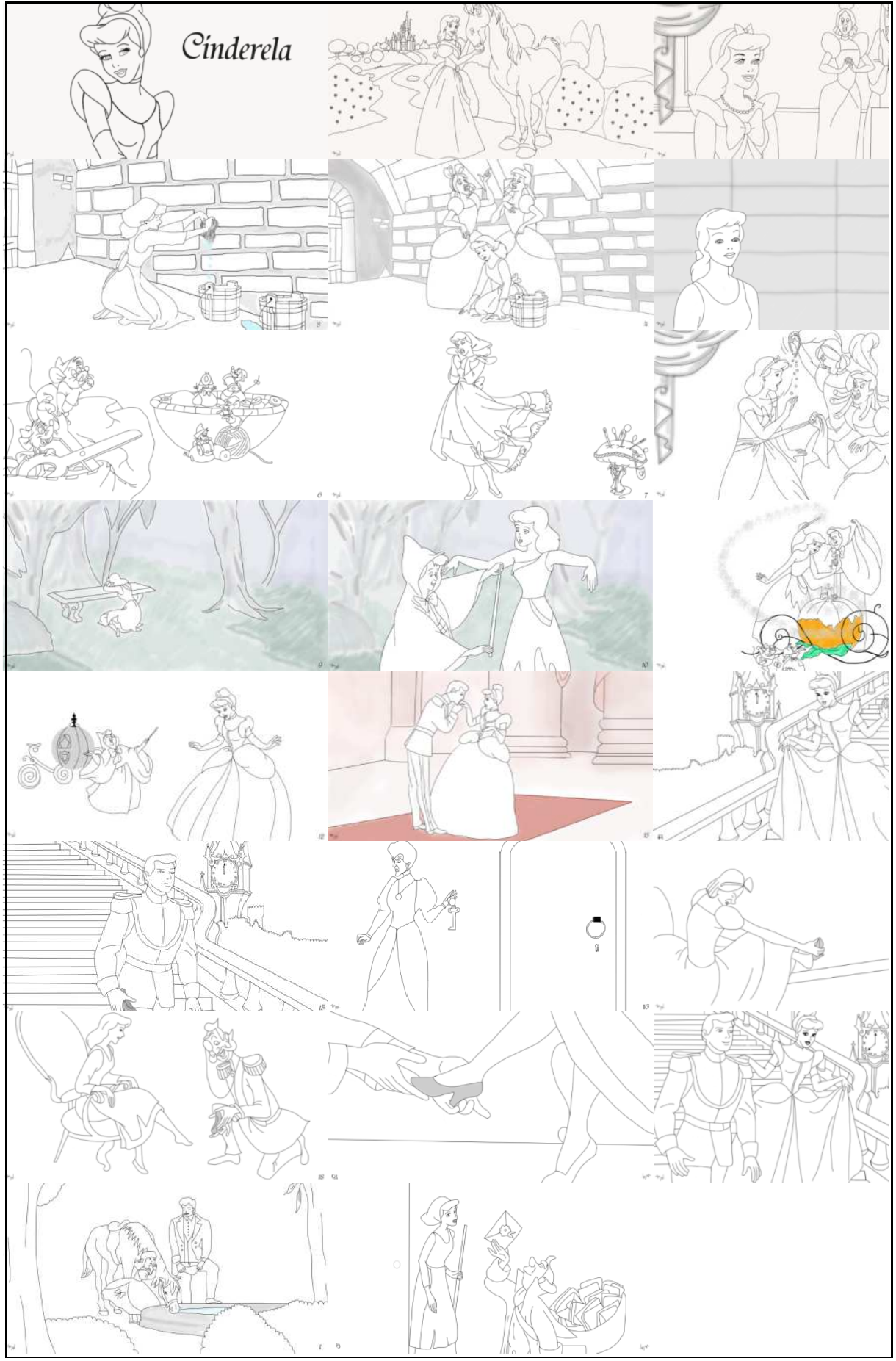}
	\caption{Sequence of Pictures of the of Cinderella story.} 
	\label{fig:book:cinderella}
\end{figure}

Figure \ref{fig:book:cinderella} is a sequence of pictures from the Cinderella story, which were used to elicit speech narratives.

\subsection{Examples of transcriptions\label{sec:supplemental:example}}

Below follows an example of a transcript of the  Cookie Theft dataset.

You just want me to start talking ? Well the little girl is asking her brother we 'll say for a cookie . Now he 's getting the cookie one for him and one for her . He unbalances the step the little stool and he 's about to fall . And the lid 's off the cookie jar . And the mother is drying the dishes abstractly so she 's left the water running in the sink and it is spilling onto the floor . And there are two there 's look like two cups and a plate on the sink and board . And that boy 's wearing shorts and the little girl is in a short skirt . And the mother has an apron on . And she 's standing at the window . The window 's opened . It must be summer or spring . And the curtains are pulled back . And they have a nice walk around their house . And there 's this nice shrubbery it appears and grass . And there 's a big picture window in the background that has the drapes pulled off . There 's a not pulled off but pulled aside . And there 's a tree in the background . And the house with the kitchen has a lot of cupboard space under the sink board and under the cabinet from which the cookie you know cookies are being removed .

Below follows an excerpt of a transcript of the Cinderella dataset.  

\textbf{Original transcript in Portuguese:}

ela morava com a madrasta as irmã né e ela era diferenciada das três era maltratada ela tinha que fazer limpeza na casa toda no castelo alias e as irmãs não faziam nada até que um dia chegou um convite do rei ele ia fazer um baile e a madrasta então é colocou que todas as filhas elas iam menos a cinderela bom como ela não tinha o vestido sapato as coisas tudo então ela mesmo teve que fazer a roupa dela começou a fazer ...

\textbf{Translation of the transcript in English:}

she lived with the stepmother the sister right and she was differentiated from the three was mistreated she had to do the cleaning in the entire house actually in the castle and the sisters didn’t do anything until one day the king’s invitation arrived he would invite everyone to a ball  and then the stepmother is said that all the daughters they would go except for cinderella well since she didn’t have a dress shoes all the things she had to make her own clothes she started to make them ...

\subsection{Coh-Metrix-Dementia metrics\label{sec:supplemental:metrics}}

\begin{enumerate}
	\item \textbf{Ambiguity}: verb ambiguity, noun ambiguity, adjective ambiguity, adverb ambiguity; 
	\item   \textbf{Anaphoras}: adjacent anaphoric references, anaphoric references; 
	\item  \textbf{Basic Counts}: Flesch index, number of word, number of sentences, number of paragraphs, words per sentence, sentences per paragraph, syllables per content word, verb incidence, noun incidence, adjective incidence, adverb incidence, pronoun incidence, content word incidence, function word incidence; 
	\item  \textbf{Connectives}: connectives incidence, additive positive connectives incidence, additive negative connectives incidence, temporal positive connectives incidence, temporal negative connectives incidence, casual positive connectives incidence, casual negative connectives incidence, logical positive connectives incidence, logical negative connectives incidence; 
	\item  \textbf{Co-reference Measures}: adjacent argument overlap, argument overlap, adjacent stem overlap, stem overlap, adjacent content word overlap; 
	\item  \textbf{Content Word Frequencies}: Content words frequency, minimum among content words frequency; 
	\item  \textbf{Hypernyms}: Mean hypernyms per verb; 
	\item  \textbf{Logic Operators}: Logic operators incidence, and incidence, or incidence, if incidence, negation incidence; 
	\item  \textbf{Latent Semantic Analysis (LSA)}: 
	Average and standard deviation similarity between pairs of adjacent sentences in the text, Average and standard deviation similarity between all sentence pairs in the text, Average and standard deviation similarity between pairs of adjacent paragraphs in the text, Givenness average and standard deviation of each sentence in the text; 
	\item  \textbf{Semantic Density}: content density; 
	\item   \textbf{Syntactical Complexity}: only cross entropy; 
	\item  \textbf{Tokens}: personal pronouns incidence, type-token ratio, Brunet index, Honoré Statistics.
\end{enumerate}

\end{document}